\documentclass[lettersize,journal]{IEEEtran}
\usepackage{cite}
\usepackage{amsmath,amssymb,amsfonts}
\usepackage{algorithmic}
\usepackage{graphicx}
\usepackage{textcomp}
\usepackage{wrapfig}
\usepackage{algorithm}
\usepackage{array, multirow}
\usepackage{subcaption} 
\usepackage{stfloats}
\usepackage{url}
\usepackage{verbatim}
\usepackage{pdflscape}
\usepackage{enumitem}   
\usepackage[table,xcdraw]{xcolor}
 
\newcommand{\degree}[1]{${#1}^\circ$}

\begin{document}
\title{Evaluation Framework for Sensor Configuration Impact on Deep Learning-Based Perception}
\author{\textit{A. Gamage, V. Donzella} \\
University of Warwick\\
\thanks{A. Gamage and V. Donzella
are with the Warwick Manufacturing Group (WMG), The University of
Warwick, CV4 7AL Coventry, U.K. (e-mail: asha.gamage@warwick.ac.uk;
v.donzella@warwick.ac.uk). }}

\maketitle

\section*{\textbf{Abstract}}
Current research on automotive perception systems predominantly focusses on either improving the performance of sensor technology or enhancing the perception functions in isolation. 
High-level perception functions are increasingly based on deep learning (DL) models due to their improved performance and generalisability compared to traditional algorithms.
Despite the vital need to evaluate the performance of DL-based perception functions under real-world conditions using onboard sensor inputs, there is a lack of frameworks to implement such systematic evaluations. 
\\
This paper presents a versatile framework to evaluate the impact of perception sensor modalities and parameter settings on DL-based perception functions. Using a simulation environment, the framework facilitates sensor modality selection and parameter tuning under different operational design domain conditions. Its effectiveness is demonstrated through a case study involving a state-of-the-art surround trajectory prediction model, highlighting performance differences across the sensor modalities radar and camera. Different settings for the parameter, horizontal field of view (HFOV) were evaluated to identify the optimal configuration. The results indicate that a radar sensor with a narrow HFOV is the most suitable configuration for the evaluated perception algorithm. The proposed framework offers a holistic approach to the design of the perception sensor suite, significantly contributing to the development of robust perception systems for automated driving systems.

\section*{\textbf{Keywords}}
Automated Vehicles, Perception Sensors, Sensor Optimisation, Simulation, Trajectory Prediction

\section{\textbf{Introduction}}\label{sec:introduction}

Deciding the optimal perception sensor suite for an automated driving system (ADS) remains an important, however, highly challenging task for the vehicle manufacturers. This is due to the dependency of most ADS perception functions on the availability of complete inputs for error-free operation. Most state-of-the-art (SOTA) high-level perception functions such as manoeuvre classification and trajectory prediction are increasingly based on deep learning (DL) models \cite{RN58, RN59, RN48}. These models are often developed using complete ground truth datasets. 
The inference performance of a DL model has an inherent bias towards the scope, quality and completeness of its development dataset \cite{RN225, RN226}.
Therefore, the performance integrity of DL-based perception functions when operating with incomplete input data from onboard perception sensors remains a crucial gap to be explored for both safe and efficient operation of ADSs.

Although the outputs of high-level, DL-based perception algorithms directly impact the safe operation of the ADS functions, research to evaluate the impact of onboard perception sensor configurations on the functional performance is rare \cite{RN47, RN210, RN44}. The likely reason is the absence of suitable data to facilitate such studies since similar data logged using different sensor modalities are not readily available in most public automotive datasets \cite{RN182, RN223}.  In addition, the parameter configurations of the perception sensors remain fixed during the data collection of naturalistic automotive datasets.

Some common approaches to determine the perception sensor suite for automated  vehicles include the following \cite{RN191, RN208, RN210, RN183}. 

\begin{enumerate}[label=(\roman*)]
\item{	Base the sensor selection on empirical knowledge and prior engineering experience.}
\item{	Select the sensors as specified by the ADS} feature developers and/ or Tier 1 suppliers.
\item{  Statistical approaches based on existing sensor implementations on autonomous vehicle platforms.}
\end{enumerate}

Such initial sensor suite recommendations are then subjected to comprehensive simulation studies focused on architectural feasibility, safety-critical scenario coverage, and compliance with regulatory and packaging constraints. However, they lack the inherent link to the performance impact on the perception functions. Hence, these approaches rarely support the original equipment manufactures (OEMs) to select the most effective and cost-efficient sensors for the optimal performance of the DL-based perception functions deployed in a vehicle.

\begin{figure*}[!t]
\centering
\includegraphics[width=0.8\textwidth]{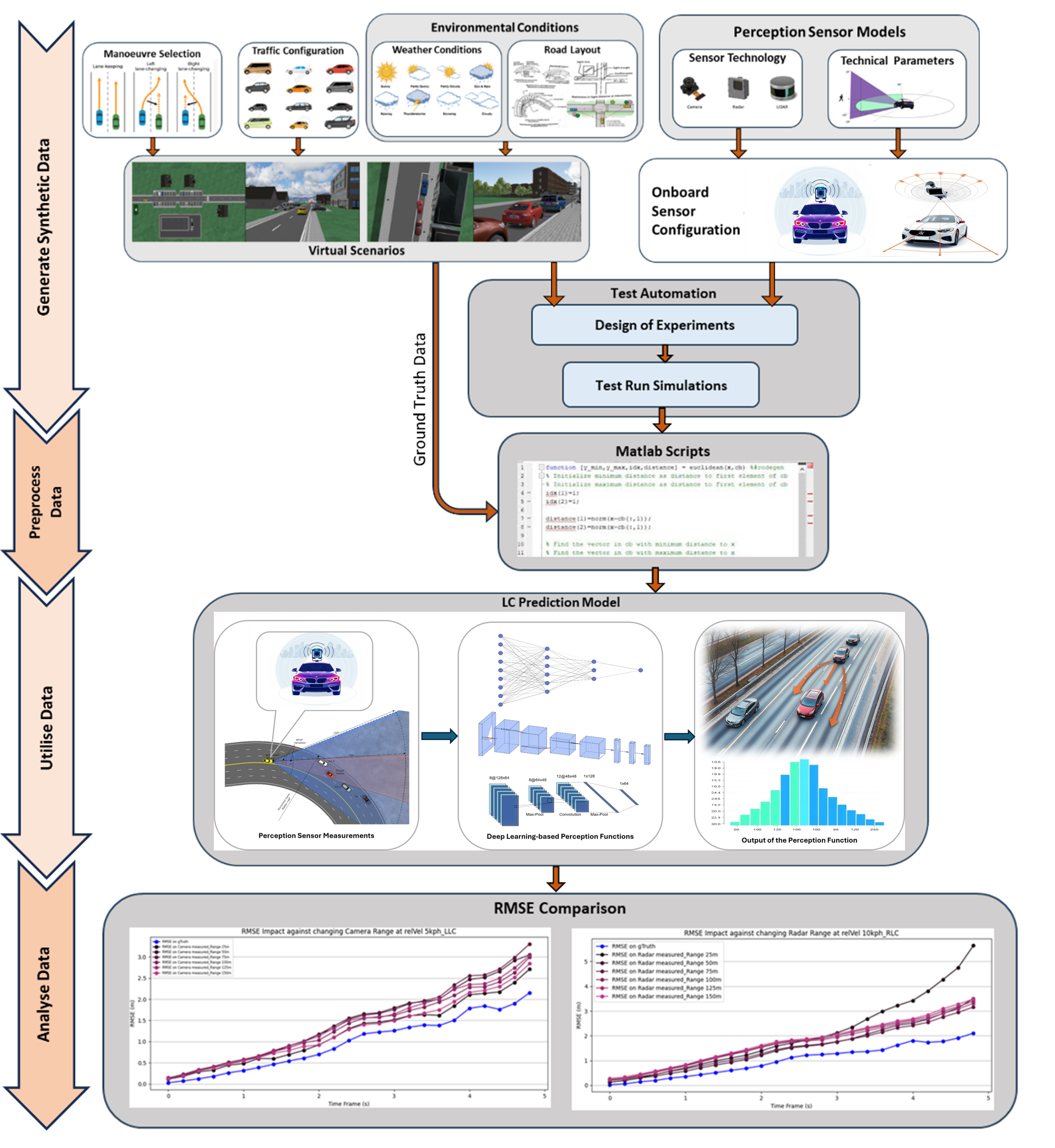}
\caption{Proposed Framework}
\label{fig:Framework}
\end{figure*}

To address this knowledge gap, a framework that facilitates exploring the impact of different perception sensor configurations on the performance of a given DL-based perception function is proposed. Developed using a virtual simulation environment, it offers maximum flexibility by adopting a modular structure, whereby virtual scenarios can incorporate a multitude of options. These include different sensor types, parameter settings, manoeuvers, changes to the environmental conditions such as weather changes and road geometries as illustrated in Fig. \ref{fig:Framework}. In doing so, the framework allows users to create scenarios that represent different operational design domains (ODDs) relevant to the perception function under evaluation, ensuring that sensor recommendations are aligned with the applicable ODDs.

The framework assumes the sensor is fixed at its standard location in current automated vehicles although the simulation tool allows for sensor placement at different positions on the vehicle body\cite{RN227}. The perception sensor configuration can be regarded as technology-agnostic, however, for generating synthetic data to conduct case studies, sensor models  from currently available sensors are used.

Existing literature on sensor optimisation only focuses on selecting and positioning sensors to optimise basic perception tasks, such as detection and classification. However, the proposed framework targets high-level DL-based perception functions at or beyond SAE Level 4 \cite{RN188}, where proactive planning replaces reactive planning and control, used in driver assistance systems such as Cruise Control (ACC) and Lane Keep Assist (LKA).  

For DL-based perception models, the impact of sensor data quality on model performance is non-deterministic, making the framework to adopt an experimental approach. Therefore, it does not aim to optimise the entire sensor suite for an automated vehicle at once but serves as a tool to iteratively identify the most suitable sensor configurations for the each deployed DL perception function. Such knowledge enables the OEMs to make informed decisions on the overall sensor suite based on the deployed perception functions.

To demonstrate the utility of the proposed framework, the impact of radar and camera sensors is evaluated on a DL-based trajectory prediction model and the optimal setting for a common sensor parameter is determined. Although the case study focuses on surrounding vehicle trajectory prediction, the proposed framework is applicable to other high-level perception functions such as manoeuvre classification and situational awareness.

The contributions of this paper are as follows.

\begin{itemize}
\item{Presents a novel framework to evaluate the impact of different perception sensor configurations on the} performance of DL-based perception functions.
\item{Offers a data-driven approach to identify }the optimal perception sensor modalities and parameter configurations for an ADS function at the early stages of vehicle development.
\item{Demonstrates the use of the proposed framework via a case study, which evaluates the performance impact of different sensor modalities and parameter settings on a SOTA trajectory prediction model.}
\end{itemize}

In addition, the `synthetic data generation' stage of the framework can enhance the diversity of existing real-world datasets by augmenting them with synthetic data  to introduce edge cases or to compensate for missing data. Such extended datasets can facilitate retraining models to uncouple domain biases, resulting in robust overall performances \cite{RN225, RN226}.

\section{\textbf{Related Works}}

Automated vehicles tend to have several established perception sensor modalities onboard such as radar, LiDAR, and vision sensors \cite{RN115, RN148, RN210, RN227}. The number of sensors and sensor modalities on a vehicle seem to increase in conjunction with the ascending levels of automation defined by the Society of Automotive Engineers (SAE)\cite{RN188, RN210}. 
Research on optimising an ADS perception sensor suite with or without linking to the performance of the perception algorithms is scarce\cite{RN208, RN217}. It involves considering different perception sensor technologies and parameter settings such as range, resolution, and field of view (FOV), alongside factors including the environmental conditions applicable to the relevant ODD.

Most of the existing studies  on automotive perception sensor optimisation define the problem as the optimal placement of perception sensors to ensure maximum coverage of a predefined area/  volume. Additionally, they only concentrate on exploring a single sensor modality as opposed to comparing different modalities \cite{RN201, RN217, RN199, RN224}. These approaches also lack the cost-benefit analysis of how sensor data impact the performance of targeted perception functions \cite{RN199, RN201, RN224}. 

A detailed survey by Kiraz \textit{et al.} provides a structured overview on the selection and positioning of sensors based on two categories; onboard and off-vehicle, with the focus of achieving cooperative, connected, and automated mobility. It highlights the limited research on optimising onboard multi-sensor configurations \cite{RN217}.

T. Meng \textit{et al.} present a framework for automatic selection and arrangement of the perception sensors using a genetic algorithm to solve for the optimal sensor suite. The approach involves several prerequisites including the need to pre-list candidate sensor types and positions based on experience and feasibility, identify a critical point set for emphatic detection, and predefine known ‘constraints’ based on prior engineering experience of the ADS function.
This limits their approach to low SAE level features, commonly established using imperative algorithms, such as automatic park assist. Hence the method is unsuitable for DL-based perception functions due to their opaque nature \cite{RN211}. 

A framework called `VESPA' to optimise the placement and orientation of multi-modal sensors is presented in \cite{RN216}. It uses three design space exploration algorithms to generate a sensor configuration that subsequently gets evaluated from an associated group of performance metrics. However, the optimisation is limited to SAE Level 2 autonomy features.

In \cite{RN37}, the authors introduce a simulation-based evaluation method to optimise the perception sensor positioning setup and demonstrate the same using two virtual driving scenarios to analyse the impact of radar setups on low-level perception functions. 
The use of a development tool, Baselabs Create to synthesise sensor data from the ground truths limits the method's ability to evaluate a sensor based on varying parameter settings.  
In a follow-up publication, the authors apply their methodology to explore radar sensor positioning, specifically by altering the mounting heights to maximise the surround view coverage \cite{RN38}. 

In \cite{RN36}, the authors evaluate the perception performance on lane recognition as a function of the sensor data quality. However, the data quality is contingent upon various combinations of imaging sensors and camera lenses, and the study is constrained to evaluating the camera sensor's capability in lane recognition. recognition’.

In an interesting attempt to link the selection of sensors to the performance of the perception function, T. Ma \textit{et al.} introduce a novel metric to evaluate the sensor configuration of an ADS.  
The evaluation metric, ‘perception entropy’, is calculated using Bayesian possibility theory to estimate the uncertainty of the detected targets and is formulated as a Gaussian distribution, conditioned on the sensor configurations. 
The definition of the metric is linked to a selected perception algorithm, hence a change to the algorithm results in a change to the relevant sensor measurements and the algorithm’s performance \cite{RN208}. 

In general, an established technique to optimise the overall sensing capability of a vehicle is sensor fusion. Many sensor-fusion algorithms have been published to leverage the individual strengths of different sensor modalities whilst compensating for each other’s weaknesses to optimise an ADS functional performance \cite{RN194, RN45, RN206}. However, the objective of most of these algorithms is to offer enhanced coverage and availability of sensor data; therefore, they lack direct associations with the requirements of the deployed perception algorithms, often leading to sub-optimal sensor configurations. 

Another related research avenue is adaptive sensor weighting techniques based on algorithmic outputs \cite{RN16, RN15}. This is where dynamically varying levels of importance are assigned to different sensor inputs based on the performance of the perception algorithms. Although a promising approach, the technique is applied after making the design decisions for the sensor suite. It also poses challenges in terms of accurately estimating the online algorithmic performance and stabilising the rapid changes in weights assignment during real-time operation.

In our view, optimal sensor configuration covers a much broader scope than determining the positions and orientations of sensors, but selecting the appropriate sensor modalities and configuring their parameters to realise the optimal performance of the onboard perception functions. Due to the varying vehicle body shapes and the diverse options in DL-based perception algorithms for vehicle deployment, to achieve the optimal sensor suite, customised evaluations need to be performed to learn the interdependencies between algorithmic performance and sensor configurations. In addition, the perception sensors currently available in the market offer a range of specifications, often increasing in price with the increase in operational range. Therefore, the proposed framework, which offers an approach to evaluate the impact between different sensor configurations and the algorithmic performance of a perception function is a promising way to achieve value-adding sensor optimisation.

\section{\textbf{Problem Formulation} \label{sec:prbFrom}}

\begin{figure*}[!t]
\centering
\includegraphics[width=0.85\textwidth]{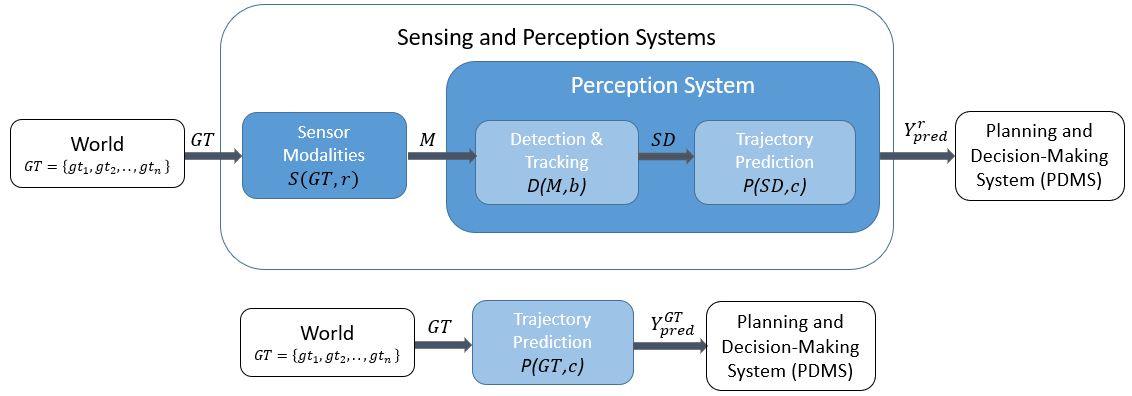}
\caption{Sensing and Perception System - Graphical Representation}
\label{fig:graphicalPipeline_2}
\end{figure*}

The evaluation framework is formulated as a modular structure that brings together the essential elements for investigating the correlations between the configuration of the perception sensors and the performance of a given ADS perception function. This section details the proposed framework using a simplified illustration of the vehicle’s sensing and perception system, specific to the function, trajectory prediction of surrounding vehicles as shown in Fig. \ref{fig:graphicalPipeline_2}.

The sensing system is a combination of automotive perception sensors and is represented as a function \(S\), with internal parameters \(r\), and real-world observations, i.e., the ground truths \(GT\). The internal sensor parameters represent the sensor specifications such as the range, the resolution, and the FOV. 

The ground truths are represented as follows.
\begin{equation}
\label{deqn_1}
GT=\ \left\{{gt}_1,{gt}_2,.....{gt}_n\right\}
\end{equation}

where ${gt}_1$ is the track of the target vehicle and ${gt}_2,.....{gt}_n$\ are all the other traffic vehicles within a predefined boundary representing the inputs for the prediction model. Similarly, if the sensor-based data is represented as $SD$,
\begin{equation}
\label{deqn_2}
SD=\ \left\{{sd}_1,{sd}_2,.....{sd}_m\right\}
\end{equation}

where ${sd}_1$ is the track of the target vehicle and ${sd}_2$\ to\ ${sd}_m$\ are the tracks of detected traffic vehicles within the perimeter by the perception sensor under study, $S$.

Note that not all traffic vehicles may get detected by a given sensor. I.e. $n\geq m$

${gt}_i\in\ {GT}\ $\ and \ ${sd}_i\in\ SD$ where ${gt}_i/{sd}_i\rightarrow\left(a,b,c,\ldots\right)$.

The parameters $a,\ b,\ c$\  can include positional data $(x,\ y,\ z)$,\ rotational data (\textit{yaw, pitch, roll}), temporal vectors of the vehicles (\textit{velocity, acceleration}) etc. depending on the input data requirements of the selected perception function.

If the duration of a vehicle track is from $t=0$ to $t=t$, since the selected trajectory prediction model for this study only requires the positional $x,\ y$ co-ordinates of the vehicles, ${gt}_i$ can be represented as follows.
\begin{equation}
\label{deqn_3}
{gt}_i=\ \left\{\left(x_i,y_i\right)_{t=0},\left(x_i,y_i\right)_{t=1},.....,\left(x_i,y_i\right)_{t=t}\right\}
\end{equation}

Similarly,
\begin{equation}
\label{deqn_4}
{sd}_i=\ \left\{\left(x_i,y_i\right)_{t=0},\left(x_i,y_i\right)_{t=1},.....,\left(x_i,y_i\right)_{t=t}\right\}
\end{equation}

As shown in Fig. \ref{fig:graphicalPipeline_2}, the sensor captures the ground truth data of a traffic object and outputs the sensor measurement, $M$, based on the sensor parameter $r$.

\begin{equation}
\label{deqn_5}
M= S\left(GT,r\right)
\end{equation}

The detection function, $D$ outputs the detected traffic, $SD$, using the sensor measurements $M$ as the input, where $b$ represents internal parameters of the detection function. Therefore, 

\begin{equation}
\label{deqn_6}
SD\ =\ D\left(M,\ b\right)
\end{equation}

By substitution from (\ref{deqn_5}),

\begin{equation}
\label{deqn_7}
SD\ =\ D\left(S\left(GT,r\right),\ b\right)
\end{equation}

Since the detection algorithm deployed in an automated vehicle is fixed, the parameter values, $b$, are also fixed for the application. Hence (\ref{deqn_7}) can be simplified to:

\begin{equation}
\label{deqn_8}
SD\ =\ \hat{D}\left(\hat{S}\left(GT,r\right)\right)
\end{equation}

Forming a composite function, $F$: 	  
\begin{equation}
\label{deqn_9}
SD\ =\ F\left(GT,\ r\right)
\end{equation}

Equation (\ref{deqn_9}) exemplifies the direct correlation between object detection and the technical parameters of a perception sensor.
Therefore, to evaluate how a sensor parameter affects the output of the trajectory prediction model, one can compare the performance (output accuracy) of the model based on the sensor data  with different sensor parameter values \((r)\) vs the performance of the model based on the ground truths.

As pictorially illustrated in Fig. \ref{fig:graphicalPipeline_2}, the trajectory prediction model outputs a probability distribution $P\left(Y\middle| X\right)$\ over the future co-ordinates of the target vehicle's predicted trajectory represented as $Y_{pred}$.

To derive a set of predictions eliminating the impact of the sensor parameters, one can use the ground truths as the input data and use the results as the baseline to compare the impact a sensor parameter has on the model's output.
If the predicted trajectory vector based on ground truths is $Y_{pred}^{GT}$, and the predicted trajectory vector based on a preset value of $r$ for the 'sensor parameter under-study' is $Y_{pred}^r$, where,
\begin{equation}
\label{deqn_10}
Y_{pred}^{GT}=\ \left[y_{t=1}^{GT},\ldots,y_{t=f}^{GT}\ \ \right]
\end{equation}
and
\begin{equation}
\label{deqn_11}
Y_{pred}^r=\ \left[y_{t=1}^r,\ldots,y_{t=f}^r\ \ \right]
\end{equation}

for a prediction time horizon of $f$ time-frames to the future, the error between the two trajectory predictions based on the ground truths $(GT)$ and the sensor detected traffic objects $(SD)$, can be shown as,
\begin{equation}
\label{deqn_12}
{E\ }_{RMSE}^r\ =\ \sqrt{\frac{1}{f}\ \sum_{i=1}^{f}\left(\ y_{t=i}^{GT}-\ y_{t=i}^r\right)^2}
\end{equation}

Where ${E\ }_{RMSE}^r$ is directly correlated to the sensor parameter setting, $r$.

Thereby determining the optimal value for each sensor parameter, $r$, which minimises the error, ${E\ }_{RMSE}^r$,\ one can optimise the configuration of the perception sensor suite.
With a specified range of values being from $i$\ to\ $j$ for given perception sensor parameter, \begin{equation}
\label{deqn_13}
{arg\ \ {\min_{r\ \in\ \left(i,j\right)}}}({{E\ }_{RMSE}^r)}
\end{equation}

The interval $\left(i,j\right)$ depends on the manufacturer's specifications of a given sensor for a selected ODD of an ADS function.

\section{\textbf{Methodology – Proposed Evaluation Framework}}

Fig. \ref{fig:Framework} illustrates the proposed framework which implements the modular structure formulated in Sec. \ref{sec:prbFrom} consisting of four stages: (i) synthetic data generation, (ii) data pre-processing, (iii) data feeding for inference, and (iv) performance evaluation

\subsection{Synthetic Data Generation}
A virtual environment is used to simulate the required driving scenarios for data generation. The use of simulation enables reproducibility and consistency across experiments, which would be very difficult to achieve with real sensor data under real driving conditions. However, the framework is designed to be both modular and robust, to accommodate real-world data as it becomes available. Following an empirical study, IPG CarMaker (CM) is chosen as the simulation tool due to its widespread use in the industry \cite{RN4}.

\begin{figure*}[!t]
\centering
\includegraphics[width=0.75\textwidth]{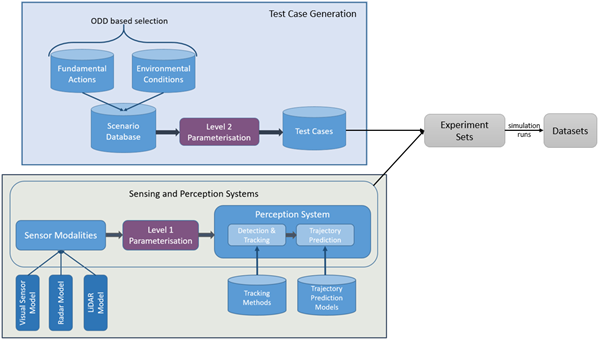}
\caption{Synthetic Data Generation Setup}
\label{fig:dataGen_Setup}
\end{figure*}

The evaluation framework provides maximum flexibility in data generation, allowing users to choose from a variety of variable settings across two avenues: (i) select the perception sensor modality and configure the parameters in line with a test strategy, (ii) select different vehicle types, manoeuvres, and environmental conditions in creating the scenarios, allowing to incorporate edge cases, as necessary. Thereby, it enables users to define scenarios that reflect various ODDs pertinent to the perception task being assessed.

Execution of the tests and data logging is automated using Test Manager, a test automation feature in CM. The logged data during simulation get saved in a results file in ERG file format.

\subsection{Data Preprocessing}

Custom data preprocessing scripts are developed in MATLAB for each sensor modality, accounting for variations in the recordable data based on the different sensor types. 
The script identifies the surrounding vehicles within a predefined boundary for the target vehicle and derives the input features for each `vehicle of interest' for the prediction model. It encapsulates the instantaneous road angle at each time frame to enable converting data between the ego-centric frame of reference and the target vehicle’s frame of reference, and generates the input data files for the trajectory prediction model. The ground truth trajectory for the target vehicle is obtained directly from the simulation tool.

The data pre-processing MATLAB scripts are available in GitHub at \underline{https://github.com/asha-gamage/perception4prediction}

\subsection{Feeding for Inference}
The ‘Utilise Data’ stage represents the use of the perceived data from the driving environment as the inputs to the prediction model to obtain inferences. Since a trajectory prediction model is used, two probability distributions of the predicted trajectory of the target vehicle are obtained, one based on the ground truths and the other based on the sensor data.

\subsection{Performance Evaluation}
The metric to use for performance evaluation is dependant on the perception function being studied to determine the optimal sensor configuration. The metric/s used during the DL model development for training and evaluation are suitable candidates. The baseline metric value needs to be established to facilitate comparison between the function's performance against parameter variations, which is based on the ground truths. Alternatively a case study specific customised metric can be defined inline with \cite{RN208}.

\section{\textbf{Case Study: Surround vehicle Trajectory Prediction}}

The merits of the framework are demonstrated using a DL-based model for trajectory prediction of surrounding vehicles. It is used to determine the appropriate sensor modality based on the HFOV and subsequently narrow down the optimal parameter value of HFOV for the function.

\subsection{Optimisation Scope and Assumptions}\label{sec:Scope}

Sensor selection is a problem based on the optimisation of a multitude of factors including the accuracy, precision, measurement resolution, range, coverage, power consumption, latency, cost, environmental robustness, etc. \cite{RN217}. However, the performance of DL-based perception models is heavily influenced by the data quality. Therefore, the optimisation basis of this study is on sensor parameters such as `range' and `HFOV' for coverage and `resolution'. Whilst acknowledging the inherent interdependencies of these sensor parameters to jointly influence the perception performance, the limitations in sensor models within simulation tools make it challenging to capture these intricate influences with full fidelity. Therefore, to simplify the experiments, a controlled experimental design is adopted to isolate the effect of a single parameter on model performance.

In addition, to isolate the impact of sensor parameter variation, sensor placement is fixed at common positions found in current automated vehicles for the two sensors, as shown in Fig.\ref{fig:senMountPos} \cite{RN210, RN217}. Since the considered trajectory prediction models are DL-based, sensitivity of the model’s performance to data quality remains unknown. Therefore, the optimisation is based on an experimental approach. 

The ODD of the trajectory prediction models in the study is highway driving, under clear daylight conditions. Therefore, the resultant sensor recommendations are applicable to similar driving environments.  In addition, the framework is specifically developed to evaluate the sensor setting's impact on the performance of high-level, DL-based perception functions. The inputs to the perception models at these levels are typically derived from object-level features such as position and velocity rather than raw sensor data. Therefore, parameters like sampling rate and resolution hardly influence the model outcome, hence not varied in this study. The outcome of the case study is only applicable to DL-based trajectory prediction deployed at or above SAE level 4.

\subsection{Model Selection}

After comparing several SOTA DL-based vehicle trajectory prediction models for their suitability for highway driving, a model proposed by N. Deo \textit{et al.} is chosen \cite{RN47}. This model predicts future vehicle motion by incorporating manoeuvre-based predictions and utilising convolutional social pooling. The manoeuvre-based approach categorises vehicle trajectories into distinct manoeuvres, whereas the convolutional social pooling helps capture spatial dependencies between the surrounding vehicles’ trajectories more effectively. The model, referred to as the CS-LSTM, combines the benefits of both the convolutional neural networks (CNN) and long short-term memory networks (LSTM) architectures and has been trained using a large naturalistic dataset, NGSIM, sourced from the US highways \cite{RN50}.

The architecture of the chosen model consists of an LSTM encoder and convolutional social pooling layers followed by an LSTM decoder.

\subsection{Inputs and outputs}

Inputs to the model are the track histories of the target vehicle and its surrounding vehicles, taken for the time frames from $t=t-t_h$  to $t=t$, where $t$ is the time when the prediction is made and $t_h$ is the start of the track history. Input features consist of the $x,\ y$ co-ordinates as detailed in \cite{RN47}. For our analysis, the track histories are derived in two ways. 

\textbf{a) Sensor-based data:} The longitudinal and lateral displacements of the detected traffic objects in the sensor frame of reference are used to calculate the track histories in the form of Cartesian co-ordinates. 
\begin{equation}
\label{deqn_14}
X^{sen}=\ \left[x_{t-t_h}^{sen},\ldots,x_t^{sen}\ \ \right]
\end{equation}

where,
\begin{equation}
\label{deqn_15}
\resizebox{.87\hsize}{!}{$x_t^{sen}=\ \left[{\hat{x}}_t^{tar},\ {\hat{y}}_t^{tar},\ {\widehat{\ x}}_t^{sur_1},\ {\widehat{\ y}}_t^{sur_1},\ldots,{\widehat{\ x}}_t^{sur_n},\ {\widehat{\ y}}_t^{sur_n}\ \right]$}
\end{equation}

are the $x,\ y$ co-ordinates of the target vehicle $(tar)$ and $n$ number of surrounding vehicles $(sur_n)$ at time frame, $t$.

b) \textbf{Ground truth data:} The ground truth Cartesian co-ordinates for the traffic objects are directly read from the simulation platform as the global positions of each traffic object. 
\begin{equation}
\label{deqn_16}
X^{GT}=\ \left[x_{t-t_h}^{GT},\ldots,x_t^{GT}\ \ \right]
\end{equation}

where,
\begin{equation}
\label{deqn_17}
\resizebox{.87\hsize}{!}{$x_t^{GT}=\ \left[{\hat{x}}_t^{tar},\ {\hat{y}}_t^{tar},\ {\widehat{\ x}}_t^{sur_1},\ {\widehat{\ y}}_t^{sur_1},\ldots,{\widehat{\ x}}_t^{sur_n},\ {\widehat{\ y}}_t^{sur_n}\ \right]$}
\end{equation}

Positional co-ordinates from both approaches are converted to the target vehicle’s frame of reference before being used as inputs to the model.
The output of the model infers the conditional probability distribution $P\left(Y\middle| X\right)$. Therefore, the inferred future $x,\ y$ positions at each future predicted time-frame are the parameters of a bivariate Gaussian probability distribution, where $\mu^t$ is the mean vector and $\Sigma^t$ is the covariance matrix.
\begin{equation}
\label{deqn_18}
\mu^t = 
\begin{pmatrix} \mu^t_x \\
 \mu^t_y \end{pmatrix}, \Sigma^t = \begin{pmatrix} ({\sigma^t_x})^2 & \sigma^t_x \sigma^t_y \rho^t\\ \\
 \sigma^t_x \sigma^t_y \rho^t & ({\sigma^t_y})^2 \end{pmatrix}
\end{equation}

The mean values are used as the inferred $x,\ y$ positional co-ordinates.

\subsection{Scenario Generation}

Six driving scenarios are created with a target vehicle performing a lane change {(LC) }manoeuvre on a highway, four representing a road segment from the US Highway 101 (replicating the real-world environment of the dataset used for training the CS-LSTM model) and the remainder a straight highway road segment. The scenarios cover the most influential positions for a LC by a target vehicle to impact the trajectory of the ego car, a LC into the same lane as the ego car, directly in front of it, and in front of the lead vehicle to the ego car.

LC is identified as one of the safety-critical manoeuvres causing a high number of road accidents on highways \cite{RN53, RN52}. Therefore, developing knowledge on the impact of perception sensor configuration on LC prediction performance is of high value.

Although some OEMs have started localising the perception sensor suite based on the country, the experiments were designed to explore bi-directional LCs since the objective is overall safe driving on highways including LCs in relation to overtaking, exiting the highways, and to accommodate offensive driving. 

\vspace{0.2cm} 
\begin{itemize}
    \item  \textbf{Sc-01}: Scenario LC to the Right - In front of the Ego car (US101)         
    \item  \textbf{Sc-02}: Scenario LC to the Left - In front of the Ego Car (US101)
    \item  \textbf{Sc-03}: Scenario LC to the Right - In front of the Lead Car (US101)
    \item  \textbf{Sc-04}: Scenario LC to the Left - In front of the Lead Car (US101)
    \item  \textbf{Sc-05}: Scenario LC to the Right – In front of the Lead Car (Straight Road)
    \item  \textbf{Sc-06}: Scenario LC to the Left – In front of the Lead Car (Straight Road)
\end{itemize}
\vspace{0.2cm} 

Detailed descriptions of Sc-01 and Sc-02 as per \cite{RN51} and simulation recordings to visualise the scenarios are available at \underline{https://github.com/asha-gamage/perception4prediction\_1}

\begin{figure}[!t]
\centering
\includegraphics[width=0.5\textwidth]{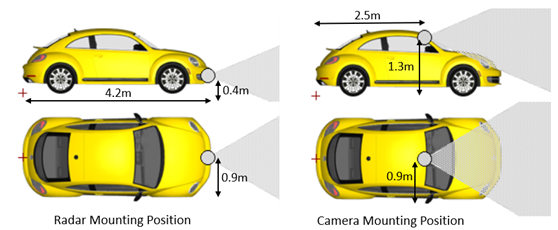}
\caption{Mounting Positions of Sensors\\ \footnotesize Source: IPG CarMaker}
\label{fig:senMountPos}
\end{figure}

\vspace{0.2cm} 
The ego car is represented by a typical passenger vehicle type along with several different vehicle types representing the surrounding vehicles. High-fidelity sensor models for radar and stereo camera from CM are mounted on the ego vehicle positioned as shown in Fig. \ref{fig:senMountPos}. The radar model uses signal-to-noise ratio for detecting the surrounding objects with consideration given to propagation and atmospheric losses whereas the camera model uses a bounding box of the object for calculating occlusions to determine whether an object is detected. The detection confidence is compared with a predefined threshold, which for this study is set to 0.6. Tables \ref{tab:radarParams} and \ref{tab:cameraParams} list the default sensor specifications for radar and camera sensors except for the technical parameter being changed, horizontal FOV (HFOV). The range of the sensor is fixed at 100 m.

 \begin{table}[!ht]
 \caption{Radar Sensor Parameters\label{tab:radarParams}}
    \centering
    \begin{tabular}{|l l l|}
    \hline
        \textbf{Technical Parameter} & \textbf{Value} & ~ \\ \hline
        ~ & \textbf{Accuracy} & \textbf{Resolution} \\ \hline
        Distance (m) & 0.4 & 1.8 \\ 
        Azimuth (deg) & 0.1 & 1.6 \\ 
        Speed (km/h) & 0.1 & 0.4 \\ 
        Cycle time (ms) & 60 & ~ \\ 
        Transmit frequency (GHz) & 77 & ~ \\ 
        Transmit Power (dBm) & 14 & ~ \\ 
        Noise bandwidth (Hz) & 25000 & ~ \\ 
        Noise figure (dB) & 4.8 & ~\\ \hline
    \end{tabular}
\end{table}

\begin{table}[!ht]
\caption{Camera Sensor Parameters\label{tab:cameraParams}}
    \centering
    \begin{tabular}{|l l l|}
    \hline
        \textbf{Technical Parameter} & \textbf{Value} & ~ \\ \hline
        ~ & \textbf{Horizontal} & \textbf{Vertical} \\ \hline
        Resolution (px) & 1280 & 960 \\  
        Resolution factor [-] & 50 & ~ \\  
        Camera type & Stereo & ~ \\  
        Baseline (m) & 0.12 & ~ \\  
        Disparity error (px) & 0.5 & ~ \\  
        Focal length (px) & 2400 & ~\\ \hline
    \end{tabular}
\end{table}

\subsection{Design of Experiments}

A test run is established using a predefined scenario with one sensor modality and the sensor's HFOV set as a variable. It is then simulated iteratively, varying the assigned value to the variable, HFOV, as listed in Table \ref{tab:var_HFoV}. Although sensor technologies are continuously improving and expanding their operational boundaries, the experiment limits the technical parameters in line with the commercially available models to date \cite{RN146, RN100}. The parameter variations are selected in step-wise increments compared to a continuous range to enable identifying the optimal sensor settings based on commercially available sensors.

To study the influence of a sensor parameter in isolation due to the ration ale in Sec. \ref{sec:Scope}, all parameters, except the one being evaluated, are kept fixed.
Identical sets of simulation test runs are conducted for the radar and camera sensors.

\subsection{Evaluation Metric}
The metric root mean squared error (RMSE) as defined in Eq. \ref{deqn_12} is used to analyse the results. The RMSEs on the predicted trajectories from the model using both ground truth data and sensor data are plotted against the prediction time horizon to compare the impact of different sensor modalities and parameter settings.

\section{\textbf{Results Interpretation and Discussion}}\label{sec:results}

\vspace{0.2cm}
\begin{table*}[]
\caption{Test Run Definitions - HFOV Variation\label{tab:var_HFoV}}
\centering
\begin{tabular}{lllllll}
\cline{1-6}
\textbf{Relative   Velocity} & \multicolumn{2}{p{5.5cm}}{\textbf{Scenario LC01: Lane   Change to the Right}} & \textbf{} & \multicolumn{2}{p{5.2cm}}{\textbf{Scenario LC02: Lane   Change to the Left}} &  \\ \cline{2-3} \cline{5-6}
                             & \multicolumn{2}{l}{Range = 100 m                     VFOV = \degree{15}}                               &           & \multicolumn{2}{l}{Range = 100 m                      VFOV = \degree{15}}                             &  \\ \cline{2-3} \cline{5-6}
                             & \textbf{HFoV}                                 & \textbf{Test ID}                                & \textbf{} & \textbf{HFoV}                                & \textbf{Test ID}                                &  \\ \cline{2-3} \cline{5-6}
$V_R$ = 5 km/h                  & \degree{30}                                         & LC01\_RelV5\_30                                 &           & \degree{30}                                        & LC02\_RelV5\_30                                 &  \\
                             & \degree{60}                                         & LC01\_RelV5\_60                                 &           & \degree{60}                                        & LC02\_RelV5\_60                                 &  \\
                             & \degree{90}                                         & LC01\_RelV5\_90                                 &           & \degree{90}                                        & LC02\_RelV5\_90                                 &  \\
                             & \degree{120}                                        & \multicolumn{2}{l}{LC01\_RelV5\_120}                        & \degree{120}                                       & LC02\_RelV5\_120                                &  \\
                             & \degree{150}                                        & \multicolumn{2}{l}{LC01\_RelV5\_150}                        & \degree{150}                                       & LC02\_RelV5\_150                                &  \\
                             & \degree{180}                                        & \multicolumn{2}{l}{LC01\_RelV5\_180}                        & \degree{180}                                       & LC02\_RelV5\_180                                &  \\ \cline{1-6}
\end{tabular}
\end{table*}

\subsubsection{\textbf{Impact of the Radar sensor's ‘HFOV’ on Trajectory Prediction}}
\begin{figure*}[h!]
\centering
\captionsetup{justification=centering}  
\captionsetup[subfigure]{labelformat=empty}  

% Row 1
\subfloat[\scriptsize{RLC}]{\includegraphics[width=3.5in]{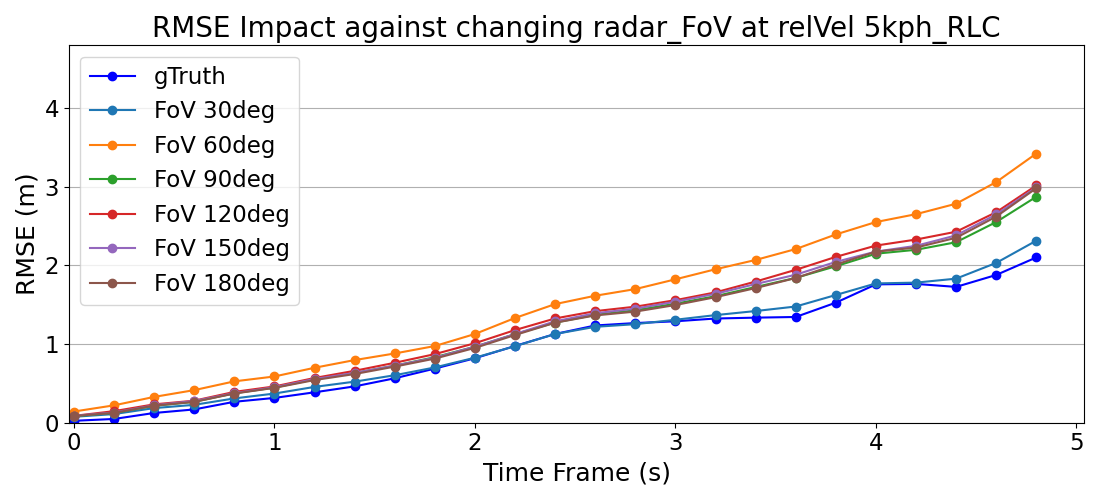}%
\label{fig_Radar_FOV_RLC- In front of Ego}}%
\hfil
\subfloat[\scriptsize{LLC}]{\includegraphics[width=3.5in]{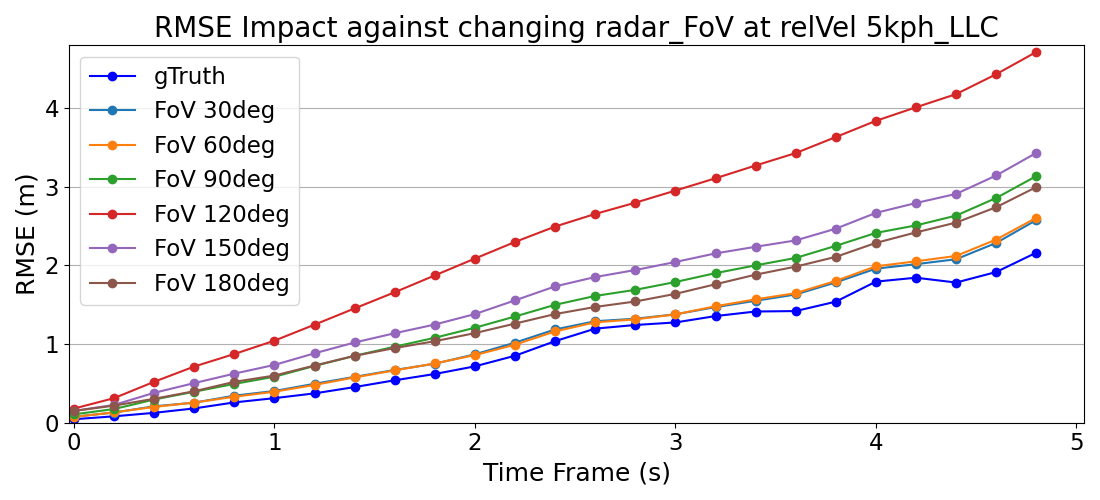}%
\label{fig_Radar_FOV - In front of Ego}} \\
\vspace{-10pt} 
\caption*{\footnotesize{Sc-01 Vs Sc-02}} 

% Row 2
\subfloat[\scriptsize{RLC}]{\includegraphics[width=3.5in]{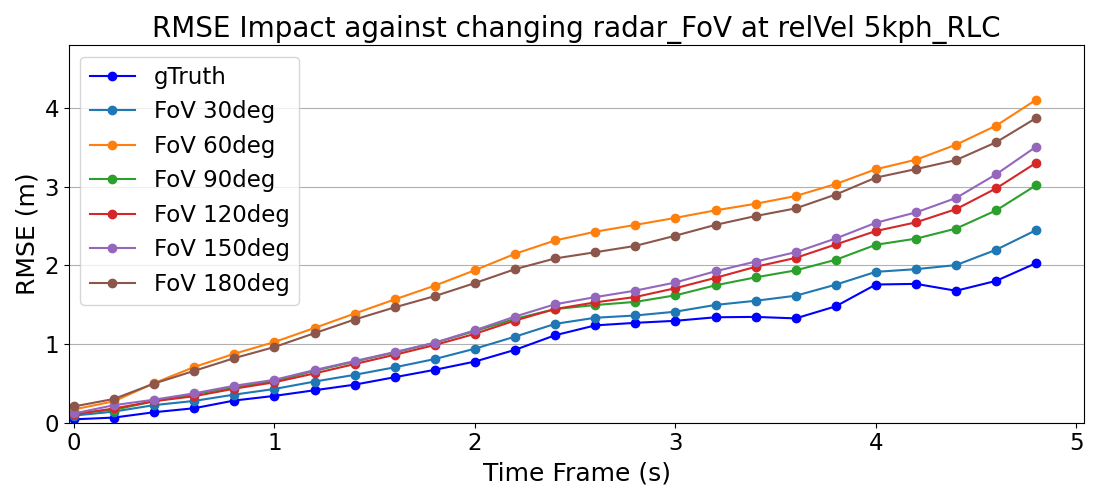}%
\label{fig_Radar_FOV_RLC- In front of Lead}}%
\hfil
\subfloat[\scriptsize{LLC}]{\includegraphics[width=3.5in]{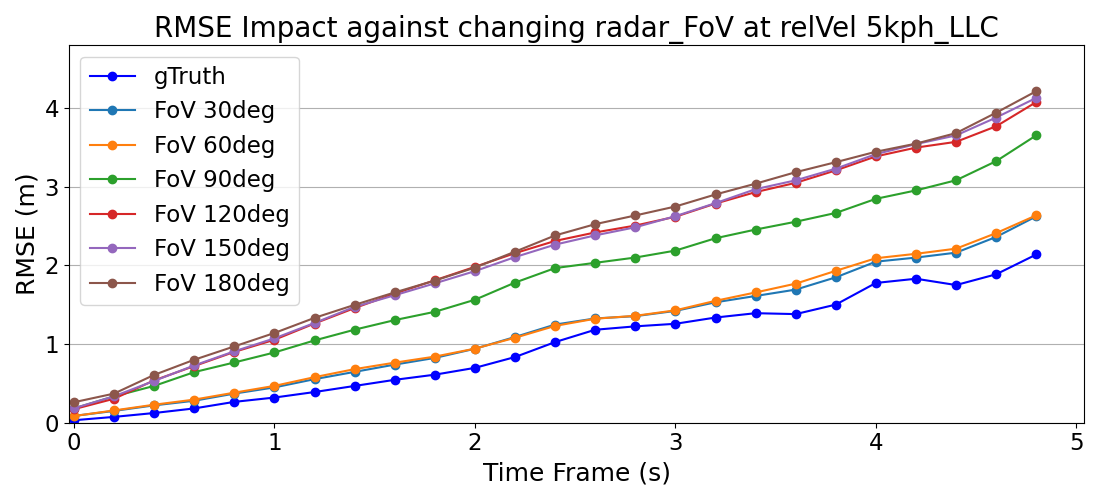}%
\label{fig_Radar_FOV - In front of Lead}} \\
\vspace{-10pt} 
\caption*{\footnotesize{Sc-03 Vs Sc-04}}  

% Row 3
\subfloat[\scriptsize{RLC}]{\includegraphics[width=3.5in]{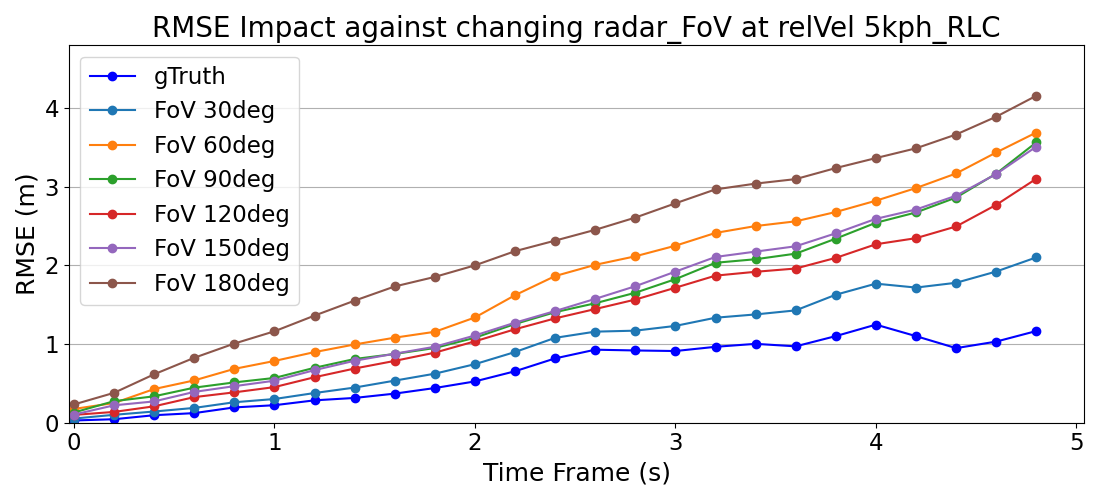}%
\label{fig_Radar_FOV_RLC- Straight Road}}
\hfil
\subfloat[\scriptsize{LLC}]{\includegraphics[width=3.5in]{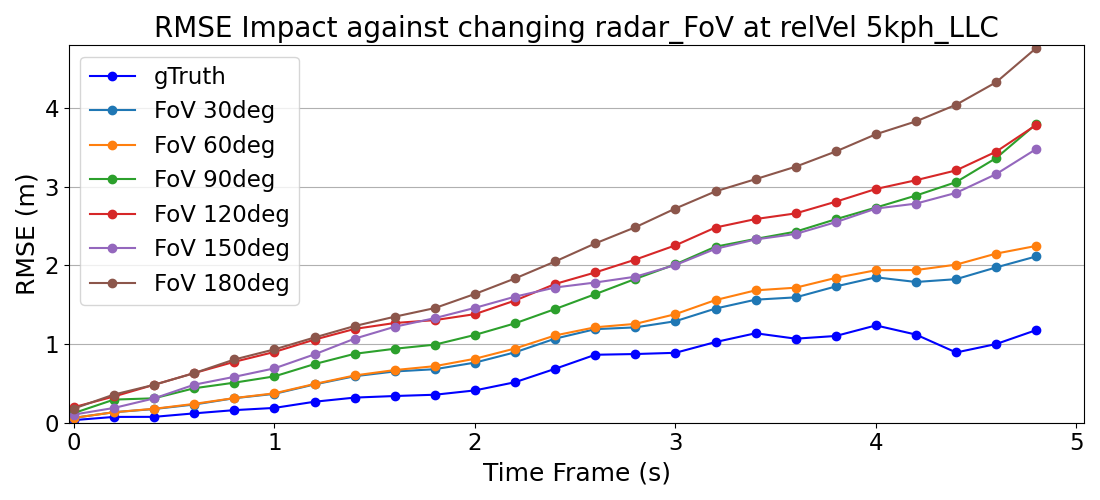}%
\label{fig_Radar_FOV - Straight Road}} \\
\vspace{-10pt} 
\caption*{\footnotesize{Sc-05 Vs Sc-06}} 

% Main Caption
\caption{Impact on RMSE - Changing Radar's HFOV: RLC Vs LLC} 
\label{fig_Radar_FOV}
\end{figure*}

\hfill\\
Fig. \ref{fig_Radar_FOV} shows the derived RMSE values based on radar data plotted against the prediction horizon of 5 s for the six scenarios. 

It is observed that the lowest HFOV setting of \degree{30} has produced the RMSEs closest to the ground truth-based RMSEs in all six scenarios explored: (i) in front of the ego car on US101, (ii) in front of the lead car on US101 and (iii) in front of the lead car on the straight road. The intuitive expectation is for the sensor-based data obtained with a larger HFOV to perform better when used as the input to the prediction model. However, the results exemplify that the RMSEs of the predicted trajectories are optimal when the measurements are taken with the lowest HFOV of \degree{30}. 

\vspace{0.3cm}
\textit{\textbf{Discussion:}}
\vspace{0.2cm}

It is believed since the dominant manoeuvre of all six scenarios is a LC into the ego car's lane, the narrow focus given by a smaller HFOV of the radar sensor enables the DL model to make a better prediction by reducing the complexity and unnecessary noise. Since for a LC, the highest impact on the target vehicle is imposed by the vehicles in its closest proximity and directly ahead, a narrow HFOV improves focus by reducing the irrelevant data of unrelated surrounding traffic. Another likely rationale is that the data captured at \degree{30} HFOV closely representing the relevant area of interest around the target vehicle. 
The CS-LSTM model uses the BEV representation of the scene with a predefined boundary around the target car. Therefore, when the sensor data is based on a wider HFOV, the inputs may appear distorted reducing the model’s performance.

The above observations highlight the significance of determining the value addition by deploying high-specification perception sensors within the sensor suite of an ADS as mentioned in \cite{RN208}. Although the expectation is for high-specification sensors leading to high-quality data, the results of this experiment indicate that the performance of DL models might not always positively correlate with data from high-specification sensors. 

\vspace{0.3cm}
\subsubsection{\textbf{Impact of the Camera sensor's ‘HFOV’ on Trajectory Prediction}}
\begin{figure*}[h!]
\centering
\captionsetup{justification=centering}  
\captionsetup[subfigure]{labelformat=empty}  

% Row 1
\subfloat[\scriptsize{RLC}]{\includegraphics[width=3.5in]{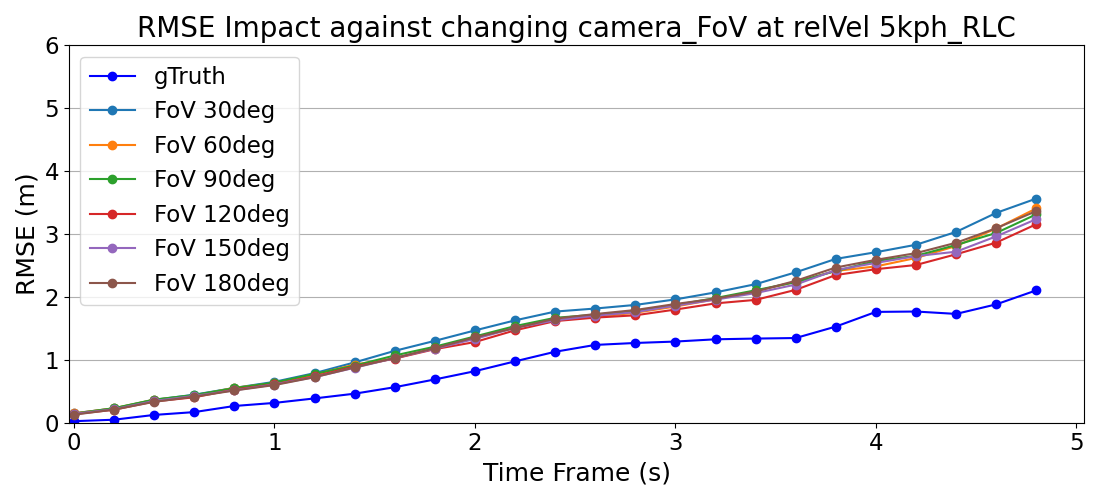}%
\label{fig_Camera_FOV_RLC- In front of Ego}}%
\hfil
\subfloat[\scriptsize{LLC}]{\includegraphics[width=3.5in]{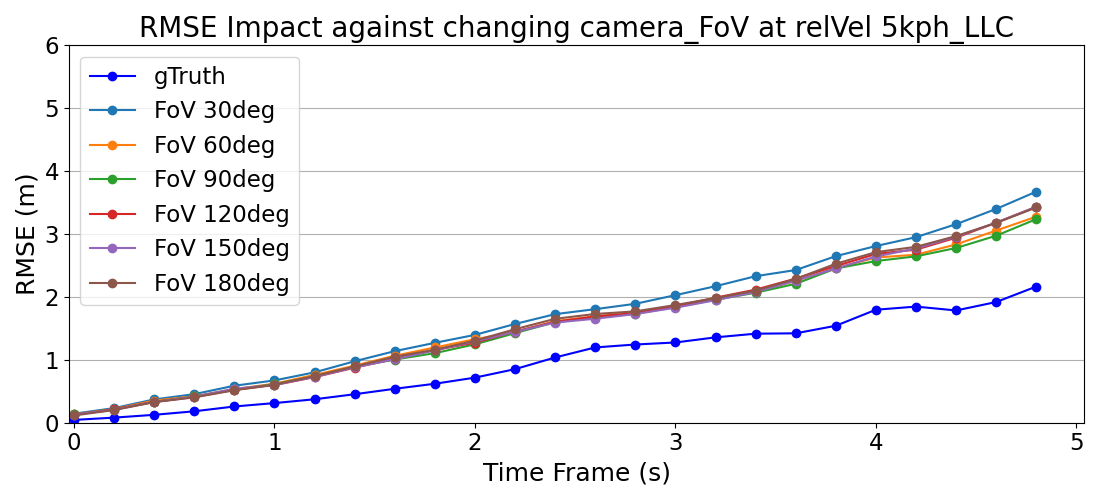}%
\label{fig_Camera_FOV_LLC- In front of Ego}} \\
\vspace{-10pt} 
\caption*{\footnotesize{Sc-01 Vs Sc-02}} 

% Row 2
\subfloat[\scriptsize{RLC}]{\includegraphics[width=3.5in]{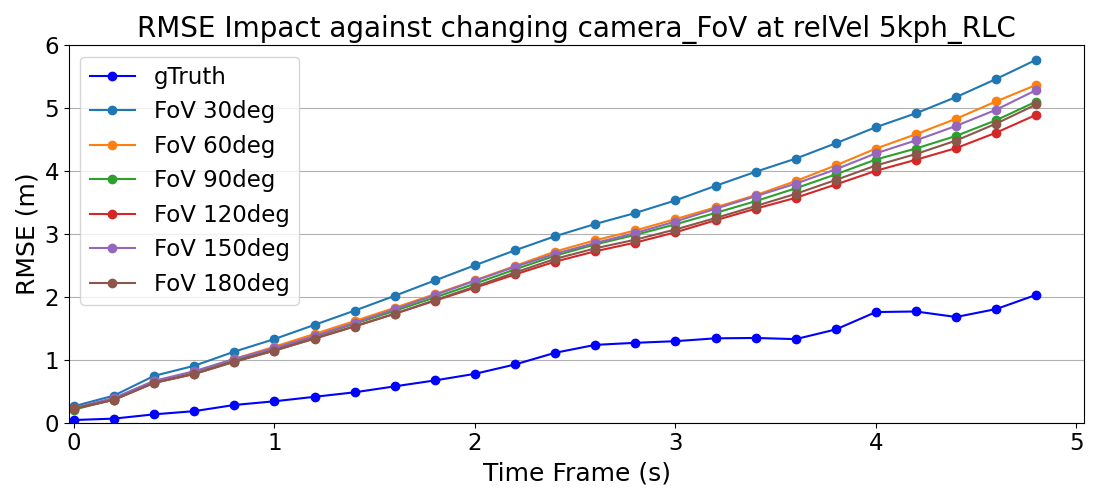}%
\label{fig_Camera_FOV_RLC- In front of Lead}}%
\hfil
\subfloat[\scriptsize{LLC}]{\includegraphics[width=3.5in]{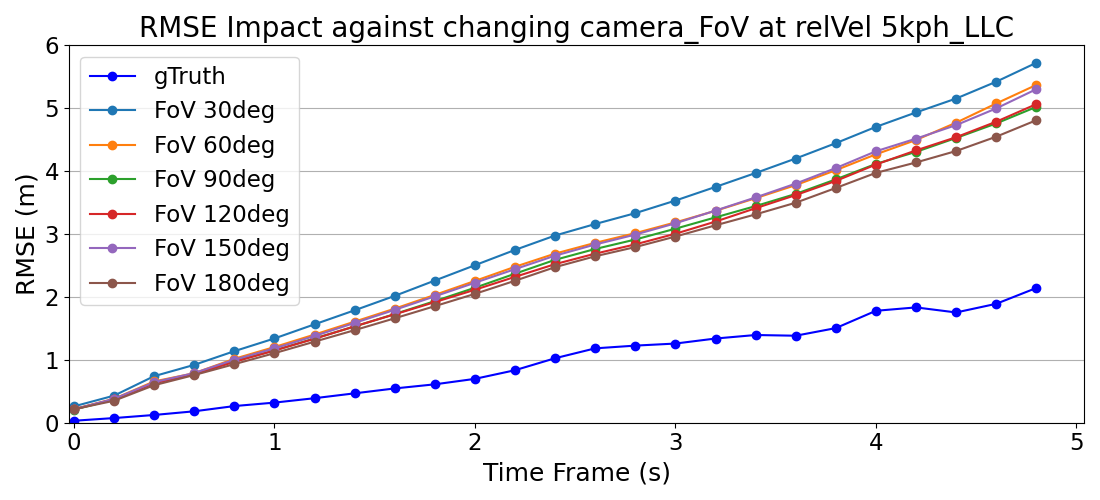}%
\label{fig_Camera_FOV_LLC- In front of Lead}} \\
\vspace{-10pt} 
\caption*{\footnotesize{Sc-03 Vs Sc-04}}  

% Row 3
\subfloat[\scriptsize{RLC}]{\includegraphics[width=3.5in]{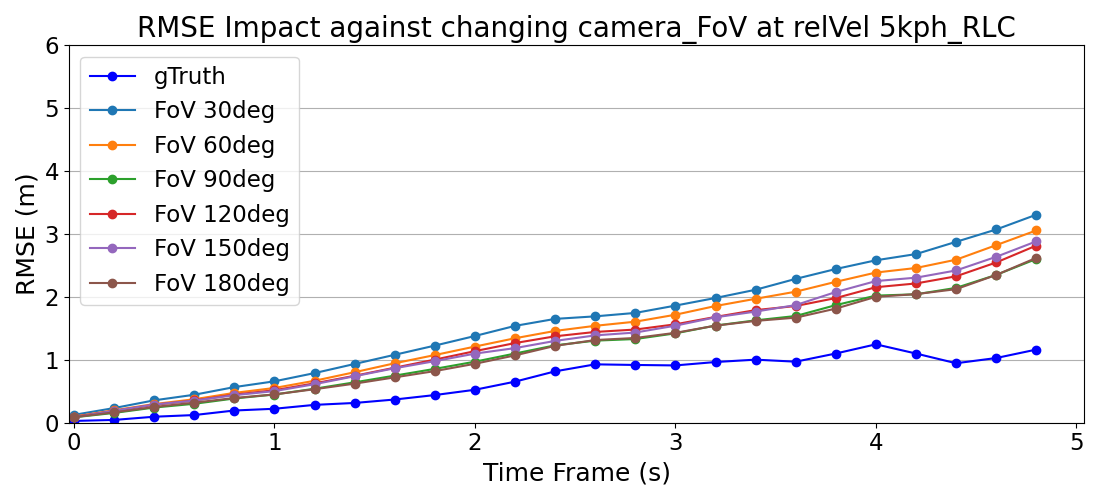}%
\label{fig_Camera_FOV_RLC- Straight Road}}
\hfil
\subfloat[\scriptsize{LLC}]{\includegraphics[width=3.5in]{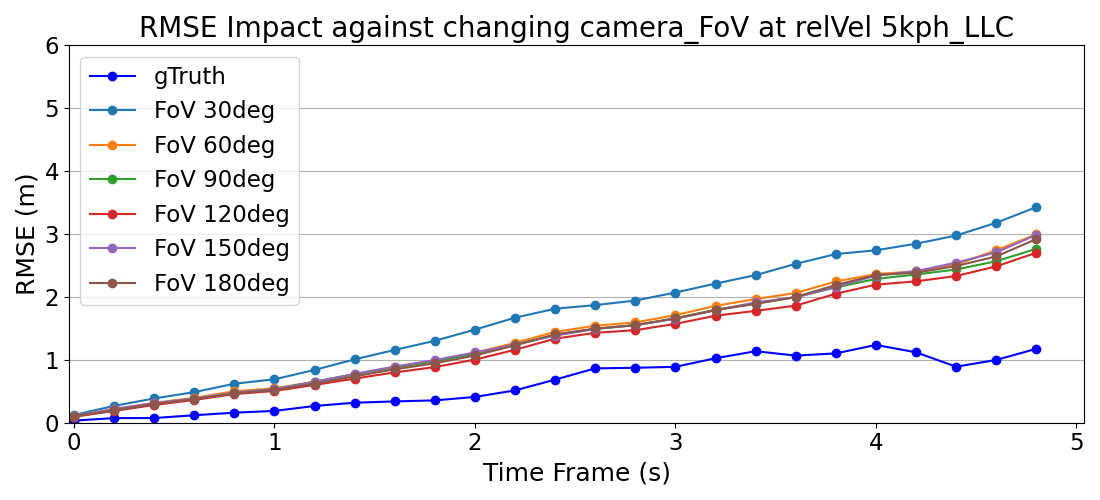}%
\label{fig_camera_FOV_LLC- Straight Road}} \\
\vspace{-10pt} 
\caption*{\footnotesize{Sc-05 Vs Sc-06}} 

% Main Caption
\caption{Impact on RMSE - Changing Camera's HFOV: RLC Vs LLC} 
\label{fig_Camera_FOV}
\end{figure*}

\hfill\\
The derived RMSE values based on camera data plotted against the prediction horizon for the six scenarios are shown in Fig. \ref{fig_Camera_FOV}.

In comparison to the observations made with radar-based data, the gaps between the camera-based RMSEs and ground truth-based RMSEs are much larger. It suggests that in general, radar is likely to be a better-suited sensor modality for surrounding vehicle trajectory prediction in comparison to camera. While further extensive data is needed to fully validate this presumption, it provides a strong foundation for further exploration.

\begin{figure}[h!]
\centering
\captionsetup{justification=centering}  
\captionsetup[subfigure]{labelformat=empty}  

% Row 1
\subfloat{\includegraphics[width=0.24\textwidth, trim= 10.5 0 10.5 0, clip]{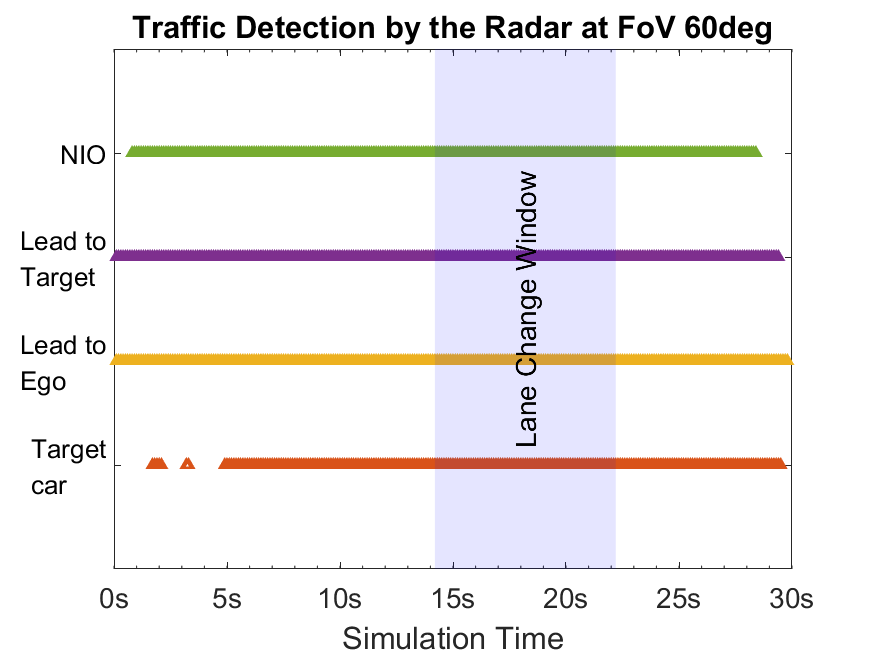}%
\label{fig_RMSEimpact by Range - CSLSTM}}%
\hfil
\subfloat{\includegraphics[width=0.24\textwidth, trim= 10.5 0 10.5 0, clip]{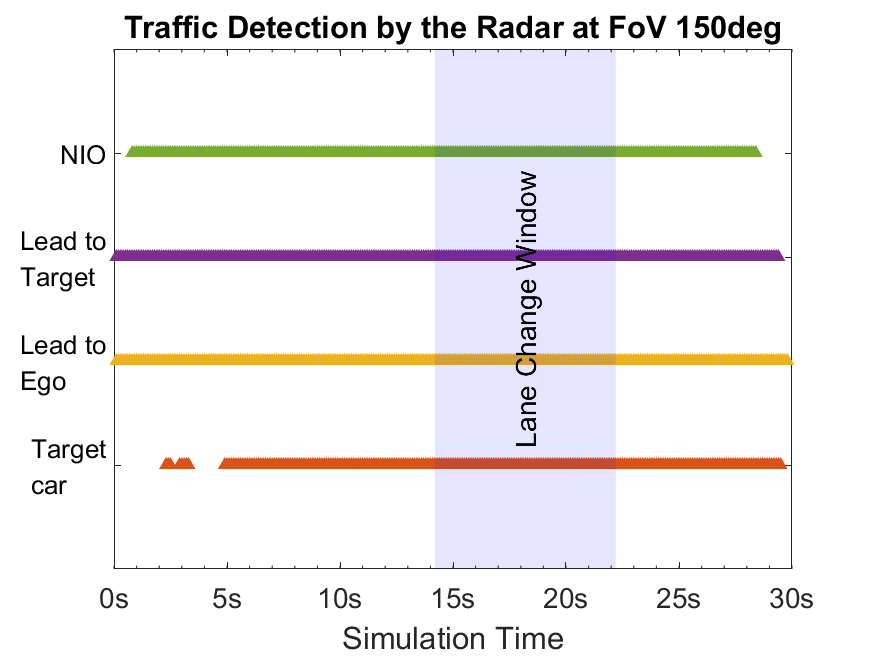}%
\label{fig_RMSEimpact by Range - STDAN}} \\
\vspace{-7pt} 
\caption*{\footnotesize{Radar Detections}} 
% Row 2
\subfloat{\includegraphics[width=0.24\textwidth, trim= 10.5 0 10.5 0, clip]{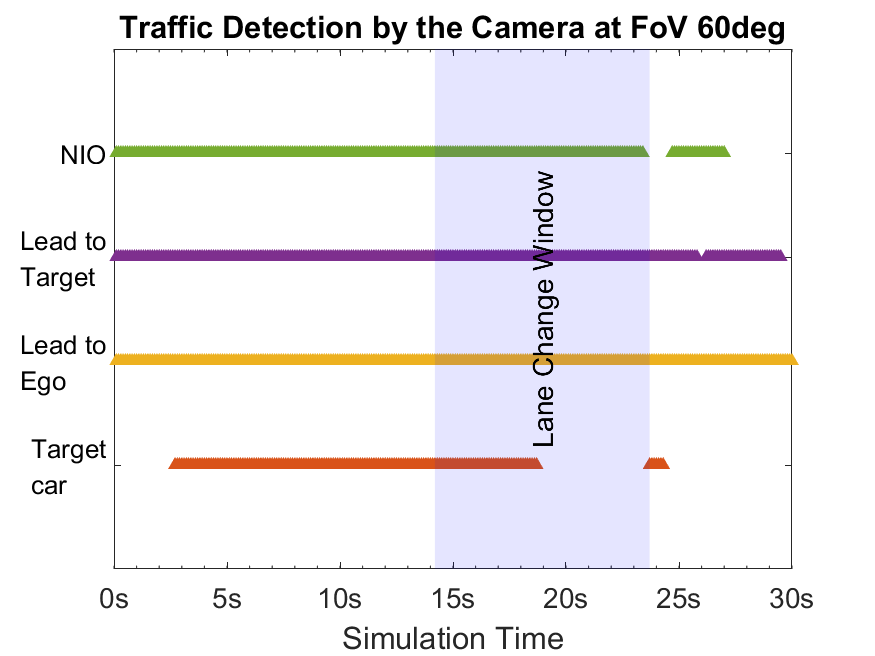}%
\label{fig_RMSEimpact by Range - CSLSTM}}%
\hfil
\subfloat{\includegraphics[width=0.24\textwidth, trim= 10.5 0 10.5 0, clip]{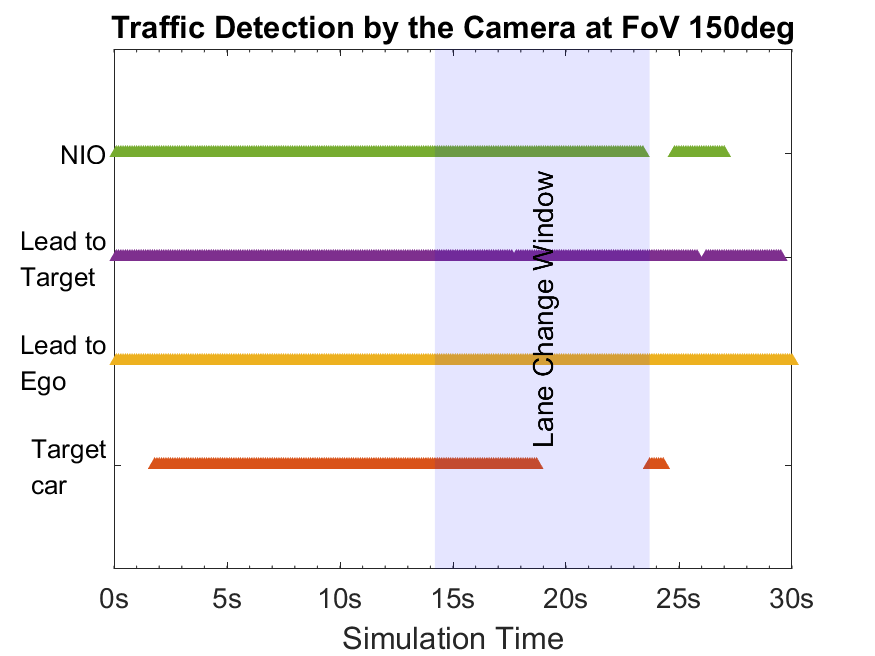}%
\label{fig_RMSEimpact by Range - STDAN}} \\
\vspace{-7pt} 
\caption*{\footnotesize{Camera Detections}} 
% Main Caption
\caption{Traffic Detection - RLC on US101 Highway} 
\label{fig_Traffic Detection on US101}
\end{figure}

\begin{figure}[h!]
\centering
\captionsetup{justification=centering}  
\captionsetup[subfigure]{labelformat=empty}  

% Row 1
\subfloat{\includegraphics[width=0.24\textwidth, trim= 9.5 0 9.5 0, clip]{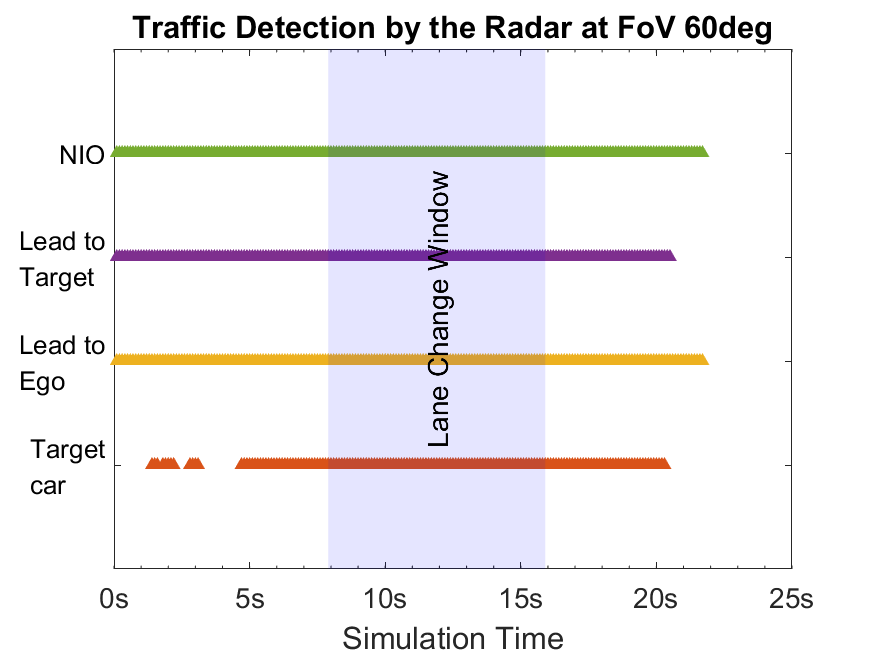}%
\label{fig_RMSEimpact by Range - CSLSTM}}%
\hfil
\subfloat{\includegraphics[width=0.24\textwidth, trim= 9.5 0 9.5 0, clip]{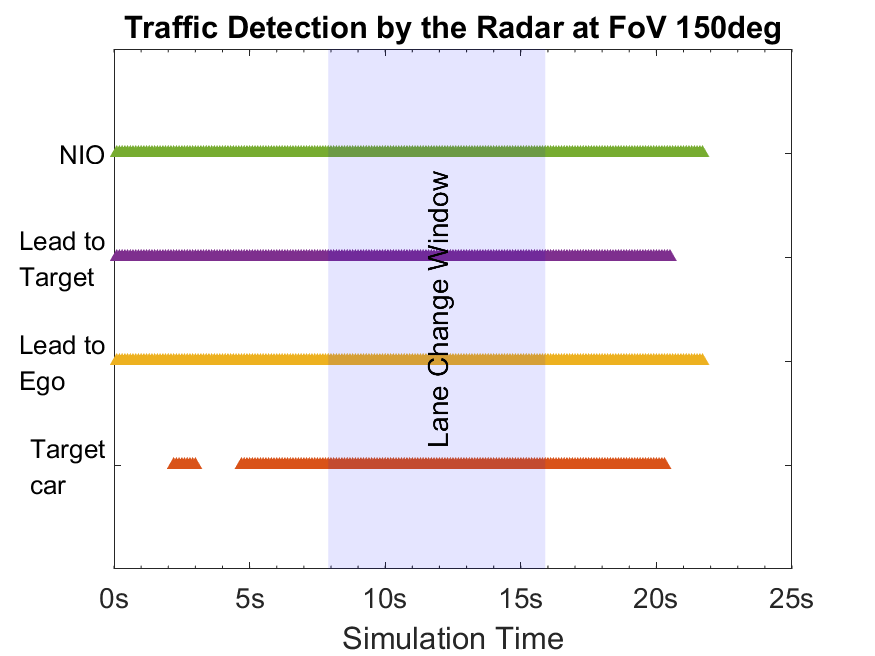}%
\label{fig_RMSEimpact by Range - STDAN}} \\
\vspace{-7pt} 
\caption*{\footnotesize{Radar Detections}} 
% Row 2
\subfloat{\includegraphics[width=0.24\textwidth, trim= 9.5 0 9.5 0, clip]{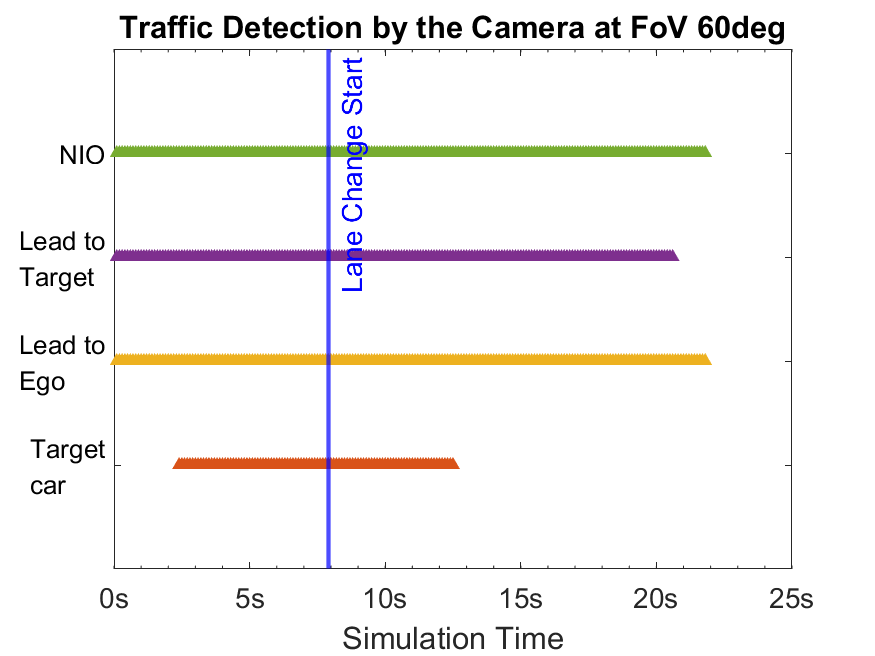}%
\label{fig_RMSEimpact by Range - CSLSTM}}%
\hfil
\subfloat{\includegraphics[width=0.24\textwidth, trim= 9.5 0 9.5 0, clip]{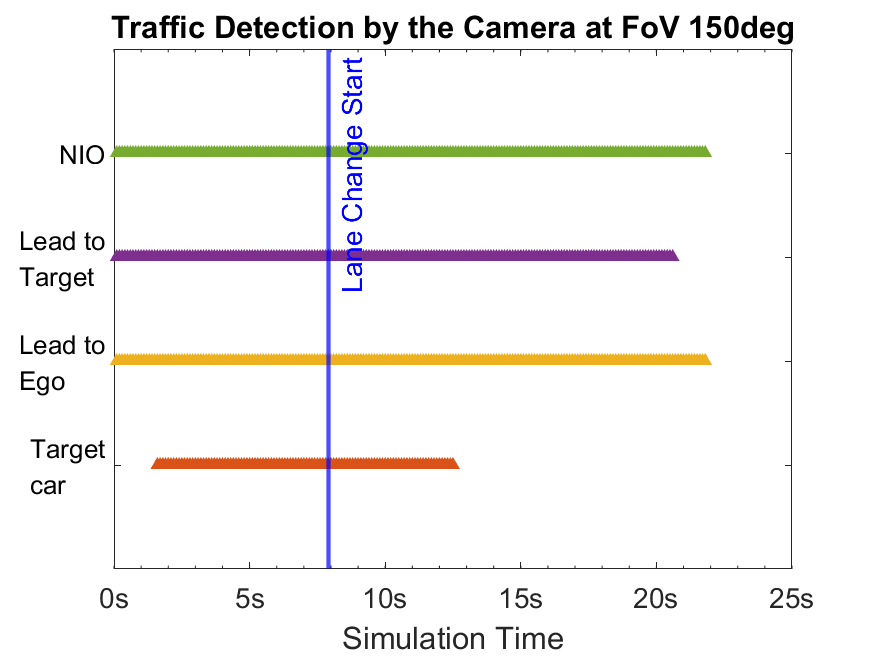}%
\label{fig_RMSEimpact by Range - STDAN}} \\
\vspace{-7pt} 
\caption*{\footnotesize{Camera Detections}} 
% Main Caption
\caption{Traffic Detection - RLC on Straight Road} 
\label{fig_Traffic Detetion on StrRd}
\end{figure}

\vspace{0.3cm}
\textit{\textbf{Discussion:}}
\vspace{0.2cm}

The camera sensor is known to struggle capturing rapid or complex movements accurately and this is likely the case for the target car and its surrounding vehicles during LCs, especially with the cut-in manoeuvres included in the scenarios. The camera is also highly susceptible to occlusions. Fig. \ref{fig_Traffic Detection on US101} shows how the occlusions leading to missing data with camera sensors erroneously elongate the duration of the LC, whereas, Fig. \ref{fig_Traffic Detetion on StrRd} shows for the straight road scenarios, the algorithm even fails to identify the end of LC manoeuvres due to the camera failing to detect the target car following the LC as it is occluded by the lead vehicle. The complete set of figures illustrating the detection of the target and the relevant surrounding vehicles between the two sensor modalities is available under Supplementary Data B at GitHub repository: \vspace{0.2cm} \underline{https://github.com/asha-gamage/perception4prediction\_1}

For LCs to the right in scenarios ‘In front of Ego’ and ‘In front of the Lead’, a HFOV setting of \degree{120} seems to provide the closest RMSEs to the ground truth RMSEs. However, for left lane changes (LLCs) in those scenarios, the HFOV settings do not show a recognisable trend in terms of producing RMSEs closest to the ground truth equivalents. The gap between the two RMSEs also remains considerably high towards the latter part of the prediction horizon. 

With Sc-05 and Sc-06, for the right lane changes (RLCs), a HFOV of \degree{180} seems to produce the RMSEs closest to the base RMSEs whereas, for the LLCs, the closest to the base RMSEs is produced with a HFOV of \degree{120}.

However, with four scenarios based on the US101 road segment, for the RLCs, a \degree{120} HFOV seems to work the best capturing all the relevant data without losing considerable resolution. However, the LLCs are inclined to be complex since they are manoeuvres into a high-speed lane and lack identifiable correlations with the HFOV changes. The considerably high RMSE gap is probably attributable to the camera’s susceptibility to dynamic scene changes and occlusions.
The discrepancy in the optimal HFOV settings between the straight road and the US101-based scenarios is likely due to the differences in the road geometry.

A consistent observation across all scenarios with the camera sensor data, is that a HFOV of \degree{30} generates the largest gap between the ground truth RMSE curve and the sensor parameter-based RMSE curve. In contrast, variations in the RMSE curves for other HFOV settings have a relatively minor impact. I.e. for camera, varying HFOV does not appear to impact the trajectory prediction model performance to a noticeable extent. 
The likely explanation is camera’s inability to capture critical inputs at the lowest HFOV of \degree{30} causing the largest gap between the RMSEs curves.

Camera seems to require a broader view to compensate for the lack of visual data, especially with the lateral movements of traffic that happen during a LC manoeuvre. On the other hand, lack of impact when increasing the camera’s HFOV beyond \degree{60} alludes to its ability to detect and process the relevant traffic objects quite robustly beyond a certain minimal HFOV,  such that further increases result in redundant data. 

Overall, the noticed difference in performance of the trajectory prediction model when the input data is sourced from two different sensor modalities is likely due to the fundamental operational differences between the two modalities. It is believed that the divergent strengths of the two sensor modalities, such as radar’s ability to focus on the relevant traffic objects with a narrow HFOV, and camera’s dependency on visual data demanding a broader view to capture the required data, causes these observations. In addition, the high sensitivity of the camera to occlusions compared to the radar’s robustness to such environmental factors is a likely factor contributing to the difference. Hence, for an ADS deploying a trajectory prediction model such as the CS-LSTM, radar becomes an essential component of the perception sensor suite due to its inherent characteristics. 

Another interesting observation is that with radar sensor inputs, the prediction model seems to perform better for the RLC manoeuvres in comparison to the LLC manoeuvres. By comparing the radar’s detections of the target vehicle and its surrounding vehicles between the bi-directional LCs, as shown in Fig. C.1 in Supplementary Data C, it is seen that the target car and the surround car, NIO, are detected less consistently in the LLC scenarios compared to the RLC scenarios. This explains the slightly better prediction performance for RLC manoeuvres, as the longer segments of missing data associated with LLC manoeuvres tend to degrade prediction accuracy.

As a crude litmus test to check the influence of missing data on the model’s prediction performance, data imputation is applied to the original radar data using linear interpolation. It is noted that with compensated data, regardless of the direction of LC, the corresponding RMSEs along the prediction horizon are compatible with each other as shown in Supplementary Data D in GitHub repository.

In all scenarios studied, the RMSEs improve following data imputation, which makes for the case that the performance of the CS-LSTM degrades considerably with impaired detection of the target car and its surrounding vehicles. This is understandable, as even short-duration detection losses directly affect the track history, which serves as the primary input to the CS-LSTM model for making predictions. This highlights the need for further research to explore the robustness of the SOTA DL trajectory prediction models when deployed with incomplete real-world data.

\section{\textbf{Conclusion}}

Insights on how different types of perception sensors and their parameter configurations contribute to optimise the perception sensor suite of an ADS are highly desirable as the automotive industry progresses towards full automation. However, studies to understand the correlations between different perception sensor modalities/ parameter settings and the performance impact on a DL-based perception functions are rare. Such knowledge is essential to optimise the overall perception system efficiently and cost-effectively. 

As a way forward, an evaluation framework is introduced which involves all stages of the perception pipeline, offering maximum flexibility with its modular approach to conduct a multitude of sensitivity studies. It facilitates quantitative sensitivity analyses between different sensor configurations and the performance of a perception function. Overall, the framework facilitates optimising the perception sensor suite by tailoring the sensor configuration to the needs of the perception algorithms implemented on the vehicle.

The contribution of our framework is demonstrated by using it to investigate the optimum parameter setting for the HFOV, and to select between the sensor modalities radar and camera for the application. The results suggest that a radar sensor with a narrow HFOV is the most suited sensor configuration for the tested perception algorithm.

The possible uses of the proposed framework to the vehicle manufacturers and the regulatory bodies include, 
\begin{enumerate}[label=(\roman*)]
        \item Making efficient design choices for the sensor suite at the early stages of the development cycle
        \item Rapid prototyping of concept vehicle platforms to make informed choices between the sensor suite and the perception algorithms
        \item Serve as a means for cost-benefit analysis on sensor technologies and sensor specifications
        \item Defining minimum safety requirements to be complied to achieve system homologation
    \end{enumerate}

Future work will extend the variety of the scenarios and sensor parameters being studied by integrating enhanced sensor models. It is also planned to explore the optimal perception sensor selections and parameter configurations for different environmental conditions such as adverse weather and challenging road geometries. 

\bibliographystyle{plain}
\bibliography{Bibliography_v3}
\end{document}